\documentclass{article}
\usepackage{spconf,amsmath,graphicx}
\usepackage{hyperref}
\usepackage{algorithmic}
\usepackage{subfigure}
\usepackage[font=small,skip=3pt]{caption}
\usepackage[linesnumbered,ruled,vlined]{algorithm2e}
\SetKwInput{KwInput}{Input}                % Set the Input
\SetKwInput{KwOutput}{Output}              % set the Output
\title{Context-Aware Change Detection With Semi-Supervised Learning}
\name{Ritu Yadav \textsuperscript{1} \thanks{The research is part of the project ‘EO-AI4GlobalChange’ funded by Digital Futures.}, Andrea Nascetti \textsuperscript{1,2}, Yifang Ban \textsuperscript{1}}
\address{\textsuperscript{1}KTH Royal Institute of Technology (Sweden), \textsuperscript{2}University of Liège (Belgium)}
\begin{document}
%\ninept
%
\maketitle
\begin{abstract}
% WHY change detection? 
% What is missing, what we proposed and achieved.
Change detection using earth observation data plays a vital role in quantifying the impact of disasters in affected areas. While data sources like Sentinel-2 provide rich optical information, they are often hindered by cloud cover, limiting their usage in disaster scenarios. However, leveraging pre-disaster optical data can offer valuable contextual information about the area such as  landcover type, vegetation cover, soil types, enabling a better understanding of the disaster's impact. In this study, we develop a model to assess the contribution of pre-disaster Sentinel-2 data in change detection tasks, focusing on disaster-affected areas. The proposed Context-Aware Change Detection Network (CACDN) utilizes a combination of pre-disaster Sentinel-2 data, pre and post-disaster Sentinel-1 data and ancillary Digital Elevation Models (DEM) data. The model is validated on flood and landslide detection and evaluated using three metrics: Area Under the Precision-Recall Curve (AUPRC), Intersection over Union (IoU), and mean IoU. The preliminary results show significant improvement (4\%, AUPRC, 3-7\% IoU, 3-6\% mean IoU) in model's change detection capabilities when incorporated with pre-disaster optical data reflecting the effectiveness of using contextual information for accurate flood and landslide detection.
% Floods and landslides are devastating natural disasters that require efficient and timely detection for effective mitigation and response. In this study we propose a deep learning model for change detection, leveraging Sentinel-1 Synthetic Aperture Radar (SAR) images, Digital Elevation Model (DEM) data, and assessing the contribution of Sentinel-2 pre-disaster images. The objective is to enhance the accuracy and reliability of detection by incorporating available contextual information from multiple sensors. We developed a Context Aware Change Detection Network (CACDN) which takes Sentinel-1 pre and post-disaster images, Sentinel-2 pre-disaster image and DEM data as input and generates a change map. Our model is evaluated on flood and landslide detection resulting in 4 to 7 \% better Intersection Over Union (IoU) when incorporated with Sentinel-2 pre-disaster images. The experimental results demonstrate the effectiveness of using contextual information in accurate flood and landslide detection.
\end{abstract}

\section{Introduction}
\label{sec:intro}
Climate change is leading to an increase in the frequency of natural hazards, with more and more people being affected by disasters~\cite{disaster_report} such as floods, landslides, wildfires and many others. An efficient evaluation of affected areas after and during a disaster event can help with quick and reliable emergency response and future action plans. Earth observations, such as multispectral and Synthetic Aperture Radar (SAR) imaging, are valuable tools for assessing and mitigating the negative impacts of natural hazards.

Sentinel satellites, such as Sentinel-2, are widely used for different classification, mapping, and monitoring tasks because of their openly accessible high spatial and temporal resolution data. Sentinel-2 is rich in information due to its multispectral imaging capabilities, which capture imagery in a wide range of visible and infrared wavelengths, providing detailed information about the earth's surface. However, most natural hazards are highly correlated with environmental conditions such as clouds, haze, and smoke, and Sentinel-2 is vulnerable to these associated artifacts.
Sentinel-1, on the other hand, is a radar satellite that acquires imagery regardless of weather or lighting conditions. Hence, primarily used in many research for monitoring changes in the earth's surface caused by natural disasters~\cite{yadav2022unsupervised, washaya2018sar, yadav2022deep}. However, the side-looking geometry and inherent characteristics of SAR images, such as speckle and layover/foreshortening, pose challenges in learning discriminative features. Consequently, SAR images generally exhibit lower classification accuracy compared to clear optical images in most common remote sensing datasets \cite{bonafilia2020sen1floods11}. Some existing works focused on combining the SAR and optical data \cite{ban2020near, 9430985, plank2016landslide}. 
Most of the disaster change detection works use post-disaster optical data which is often partially hindered by the clouds or smoke ~\cite{prudente2020sar} affecting the impact assessment capability. However, we can easily obtain clear pre-disaster optical images. Our aim is to utilize clear Sentinel-2 pre-disaster images to learn about the area context before the disaster, which can be used to improve the accuracy of post-disaster assessments. 
In this study, we propose a model that can detect changes more efficiently by learning from more contextual information. We use sentinel-1 pre and post-disaster images along with Sentinel-2 pre-disaster images, DEM and derived slope features so that model can correlate the disaster-affected area with superior spatial features and physical properties of the area.

\begin{figure*}  
\centering  
\begin{subfigure}
  \centering
  \includegraphics[width=180mm, height=88mm]{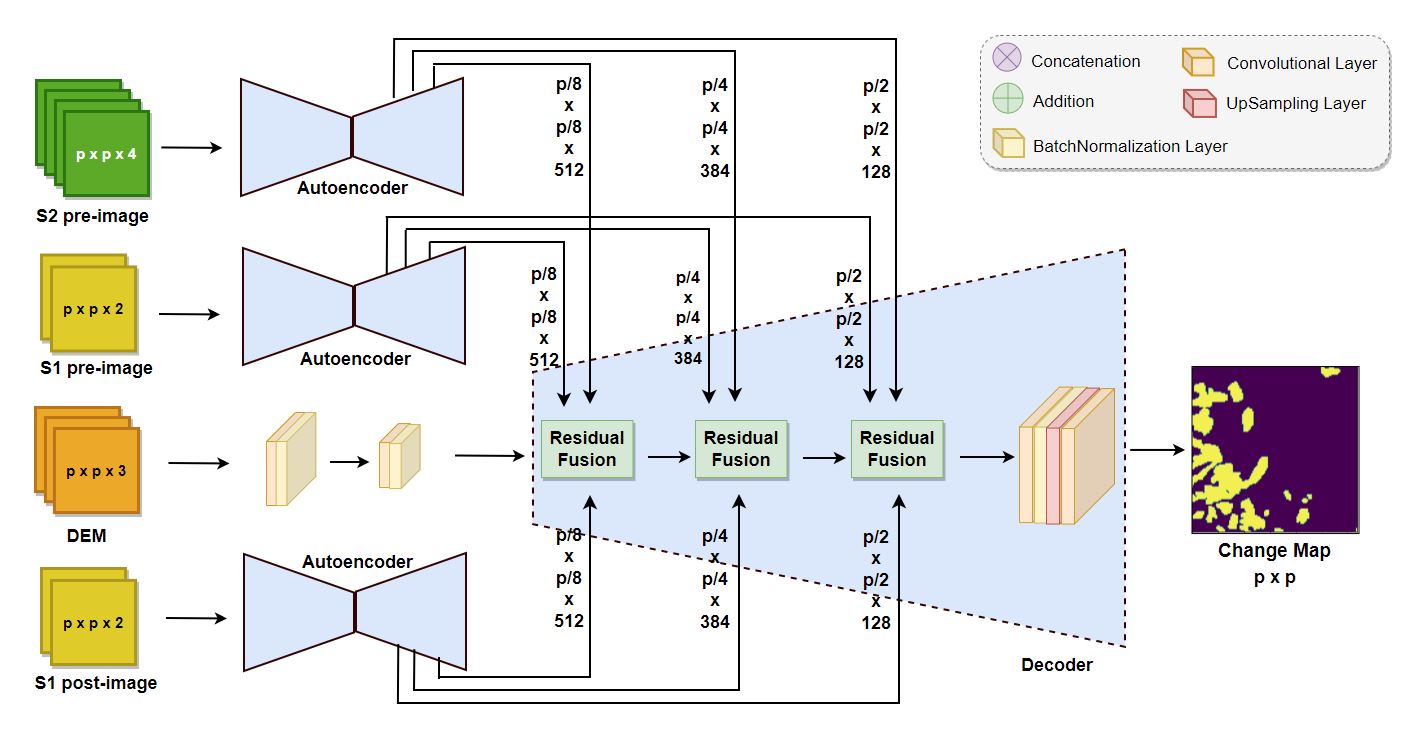}
\end{subfigure}
\caption{CACDN Architecture. The three autoencoders are pretrained on corresponding Sentinel data with self-supervision.}
\label{network}
\end{figure*}
\vspace{-13px}
\section{Data Preparation}
\label{sec:Dataset}
We tested our context-aware change detection approach on two types of disaster events namely flood detection and landslide detection. For the flood detection task, we utilized an existing dataset called Sen1floods11 \cite{bonafilia2020sen1floods11} and expanded it by incorporating additional data. The original dataset consisted of 446 Sentinel-1 post-flood patches, each measuring 512x512 pixels, along with corresponding ground truth labels. We divided each sample into smaller 256x256 pixel tiles and extended the dataset by adding Sentinel-1 Ground Range Detected (GRD) pre-flood images, cloud-free Sentinel-2 MSI Level-2A pre-flood images, DEM SRTM and it is derived data. We downloaded additional data using Google Earth Engine python API \cite{GORELICK201718}. 
The Sentinel-1 images consisted of two polarization bands (VV, VH), while Sentinel-2 had four bands (Red, Green, Blue, NIR). The DEM data provided elevation information, and we also utilized it's derived products such as slope and aspect. To maintain consistency in evaluation, we followed the data division guidelines established by the Sen1Floods11 dataset.
For the landslide detection task, we collected a small dataset encompassing two landslides one in Hokkaido, Japan (Sep 2018) and another in Mt Talakmau, Indonesia (Feb 2022). Ground truth maps for these landslides were obtained from UNOSAT. To organize the data, we divided the area of interest into tiles of size 128x128 pixels. Corresponding Sentinel-1 GRD pre-landslide images, Sentinel-1 GRD post-landslide images, Sentinel-2 MSI Level-2A pre-landslide images, DEM SRTM data, and derived products such as slope and aspect were processed and downloaded at 10m spatial resolution using Google Earth Engine python API. The dataset was split into training and test sets, with an 80-20 ratio, resulting in 227 training samples and 57 test samples.

\section{Proposed Method} 
\label{sec:method}

\subsection{CACDN Architecture}
\label{sec:Network}
\begin{figure}[htbp]
\caption{Foundation model. Autoencoder pretrained on reconstruction task.}
\centerline{\includegraphics[width=85mm, height=30mm]{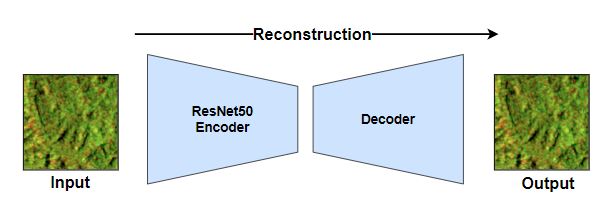}}
\label{autoenc}
\end{figure}

Figure \ref{network} shows the architecture of the proposed Context Aware Change Detection Network (CACDN), consisting of four branches to learn features separately from Sentinel-2 pre-disaster image, Sentinel-1 pre-disaster image, DEM data and Sentinel-1 post-image. The four branches take different size inputs specified network diagram where p represents patch size (width and height both). The sentinel image features are learned using autoencoders (see Figure \ref{autoenc}) as our foundation model. We used autoencoder with a Resnet50 encoder and a decoder containing four upsampling blocks (convolutional, Batchnormalization amd upsampling layer) reconstructing input. The fourth branch takes DEM input and passes through two sets of convolutional and batch normalization layers. 

\begin{figure}[htbp]
\caption{Residual Fusion block.}
\centerline{\includegraphics[width=65mm, height=43mm]{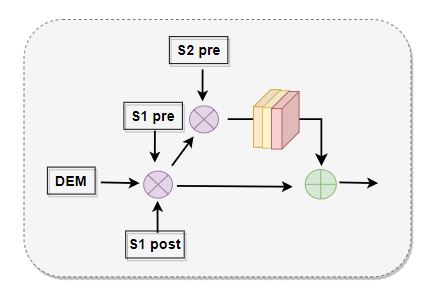}}
\label{res_fusion}
\end{figure}
\vspace{-10px}
Multi-scale feature maps from the three reconstruction networks and output DEM feature maps from the fourth branch are fused together. We propose a residual fusion technique (see Figure \ref{res_fusion}) to fuse the feature maps from the four branches, enabling model to get benefit from Sentinel-2 pre-image without introducing extra errors in training. The concept is inline with the residual networks~\cite{he2016deep} using identity blocks. The fusion takes place in three steps with different scale features (p/8 x p/8), (p/4 x p/4) and (p/2 x p/2). The fused features are then fed to a set of layers (see Figure \ref{network} for details) generating a binary change map of size (pxp).
\vspace{-5px}
\subsection{Implementation and Training}
The three autoencoders are first pretrained on reconstruction tasks with self-supervision. The reconstruction losses used are cross-entropy and mean square error with equal weight i.e. 0.5. The pretrained autoencoders are combined with the DEM processing branch and the remaining network. The full network is then trained using a weighted combination of supervised segmentation losses and self-supervised reconstruction losses given in equation Eq. \ref{eq:total_loss}. The supervised losses are focal loss ($L_{focal}$) and dice loss ($L_{dice}$). The self-supervised reconstruction losses are cross-entropy ($L_{ce}$) and mean square error $L_{mse}$).
\begin{equation} 
    \label{eq:total_loss}
    \begin{aligned}
    Loss = 0.6 * (L_{focal} + L_{dice}) + 0.4 * (L_{ce} + \beta*L_{mse})
    \end{aligned}
\end{equation}

Our proposed CACDN model is evaluated on two change detection tasks namely flood detection and landslide detection. For flood detection, we trained the model on tiles of size 256 x 256 and for landslides, we used smaller tiles 128 x128 pixels. Since our landslide data is comparatively small, we  trained a smaller version of the CACDN model. To reduce the model size we used Resnet50 encoder with three downsampling levels and decoder with three upsampling blocks. Also, the features are fused at two scales (p/4 x p/4) and (p/2 x p/2 ) with the same number of channels i.e. 384 and 128.
Training images are augmented using Gaussian blur and gamma contrast augmentation methods. We pre-trained each autoencoder for 50 epochs with batch size 4, adam optimizer and 0.0001 as the initial learning rate. The end-to-end network is trained for 300 epochs with batch size 4, adam optimizer and a smaller initial learning rate of 0.00001. For better convergence, the learning rate is decayed until 0.000001. The decay steps are controlled with the "reduce on plateau" method. 

\section{Results and Evaluation}
We evaluated CACDN model on flood detection and landslide detection using three metrics; IoU to measure overlap between the predicted change areas and the ground truth change areas, mean IoU to calculate the average IoU across all predicted and ground truth change areas, and AUPRC to assess the trade-off between precision and recall. AUPRC metric is important here because of imbalanced change detection datasets. IoU and mean IoU focus on the spatial accuracy and overlap between predicted and ground truth change areas, while AUPRC evaluates the classification performance and the balance between precision and recall.

The qualitative results on flood detection and landslide detection are given in Table \ref{SOTA_comp1} and \ref{SOTA_comp2} respectively. To show the impact of using Sentinel-2 pre-disaster images, We present a comparison of our CACDN model with Sentinel-2 pre-disaster images and CACDN model without Sentinel-2 pre-disaster images.

\begin{table}[htbp]
\caption{CACDN evaluation results on Floods test set.}
\begin{center}
\resizebox{\columnwidth}{!}{%
\begin{tabular}{l|l|l|l}
\hline
\textbf{Method} &\textbf{AUPRC} &\textbf{IoU}  &\textbf{Mean IoU} \\ \hline \hline
% DAUSAR CD Network & 0.00 & 0.700 & 0.463   \\ \hline
CACDN with S2 Pre-Image & \textbf{0.838}  & \textbf{0.738} & \textbf{0.50} \\ \hline
CACDN w/o S2 Pre-Image & 0.801  & 0.712 & 0.47 \\ \hline
\end{tabular}%
}
\vspace{-10px}
\end{center}
\label{SOTA_comp1}
\end{table}

\begin{figure*}  
\centering  
\begin{subfigure}
  \centering  
  \includegraphics[width=180mm, height=65mm]{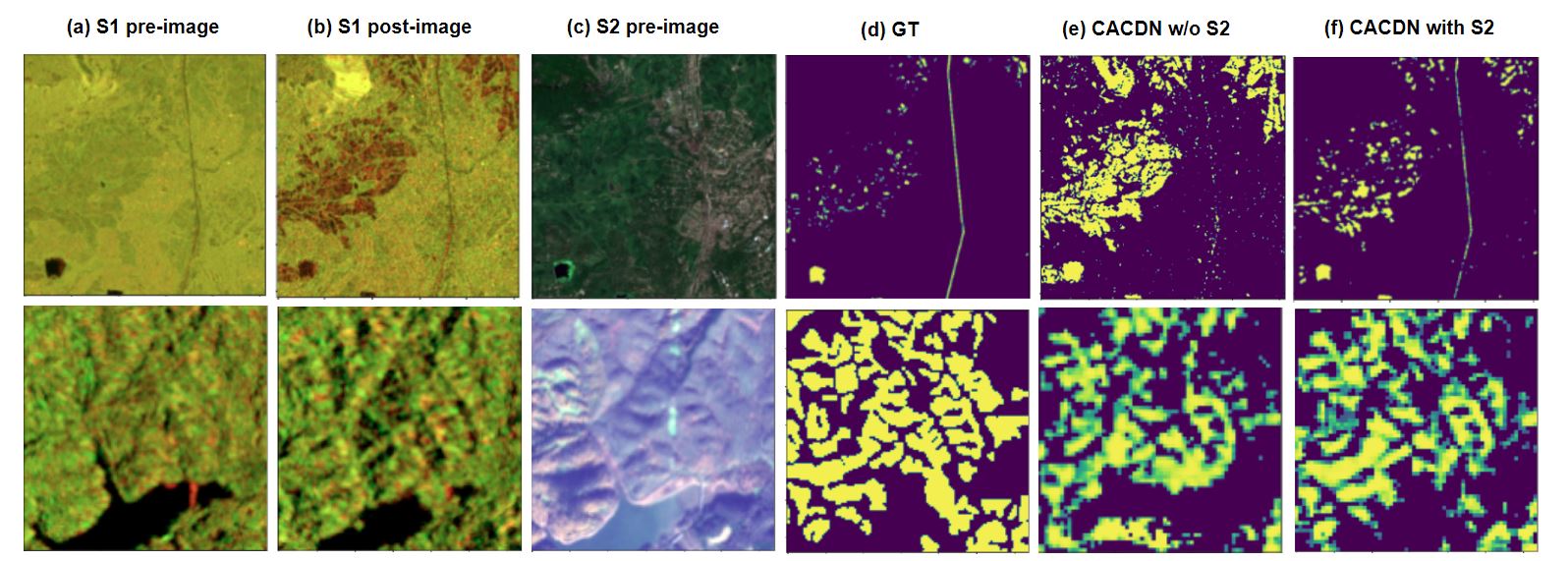}
\end{subfigure}
\vspace{-5px}
\caption{Result Samples. Two sample results are visualized in 2 rows. The first sample is from the flood test set and the second sample is from the landslide test set.}
\vspace{-10px}
\label{data_samples}
\end{figure*}

On the flood detection task, all three metric scores are good ensuring highly accurate flood detection both in terms of pixel-wise overlapping with ground truth and in terms of precision-recall balance. When trained with Sentinel-2 pre-flood image our model gave a lead of 3 to 4 \% in all three metrics (see Table \ref{SOTA_comp1}). The gain in mean IoU score indicates an improvement across test samples and a better AUPRC score shows model's ability to correctly identify positive changes. The metric scores are supported by the qualitative results in Figure \ref{data_samples}, where the model gave better detections (overlap with ground truth and low false detection rate) when trained with Sentinel-2 pre-flood image. 
On further evaluation, we see that Sentinel-2 helps to reduce confusion caused by similar features of shallow water vegetation and floods in Sentinel-1 imagery. Using Sentinel-2 pre-images, land covers are better differentiated and waterbodies, particularly narrow rivers, are better detected, leading to improved flood detection. Additionally, incorporating DEM data allows the model to learn the correlation between surface slope and flood possibility, enabling the identification of areas with high-slope surfaces that are less likely to be flooded.
\begin{table}[htbp]
\caption{CACDN evaluation results on Landslide test set.}
\begin{center}
\resizebox{\columnwidth}{!}{%
\begin{tabular}{l|l|l|l}
\hline
\textbf{Method} &\textbf{AUPRC} &\textbf{IoU}  &\textbf{Mean IoU} \\ \hline \hline
% DAUSAR CD Network & 0.682  & 0.364 & 0.274 \\ \hline
CACDN with S2 Pre-Image & \textbf{0.746}  & \textbf{0.453} & \textbf{0.363} \\ \hline
CACDN w/o S2 Pre-Image & 0.708  & 0.380 & 0.301 \\ \hline
\end{tabular}%
}
\vspace{-10px}
\end{center}
\label{SOTA_comp2}
\end{table}

Similar to the flood detection task, our model shows better landslide detection when trained with Sentinel-2 pre-landslide images (see Table \ref{SOTA_comp2}). The gain of 5 to 7 \% across three metric scores. Sentinel-1 often suffers from foreshortening and shadow effects especially in mountain areas, making it challenging to accurately detect and characterize landslides. Whereas, Sentinel-2 is less affected by the foreshortening allowing a better representation of the landscape. Furthermore, DEM and slope data provide valuable physical properties for the model to learn the correlation with the landslide event and enhance landslide detection.

\section{Conclusion}
\label{sec:conclusion}

In this study, we developed a context aware change detection model that learns contextual features from multi-source data such as surface texture features from Sentinel-1 SAR data, landcover features from Sentinel-2 MSI data, and correlation of physical properties (elevation, slope) with the type of disaster event. Since clear MSI data is not available in most of the disaster scenarios, we incorporate pre-disaster MSI data to learn from the optical features (pre-disaster) of the affected area.
We study the impact of using Sentinel-2 pre-disaster images on the model's change detection capabilities. We identified that incorporating contextual information from pre-disaster optical images helps model to learn specific characteristics and vulnerabilities of the region, leading to more accurate and reliable detection. Our model gained 4 to 7 \% IoU when training the model with Sentinel-2 pre-disaster images for flood and landslide detection tasks. 
Moving forward, our future work aims to validate our approach on larger datasets and several other types of disaster events. This will allow us to develop a more generalized model that can offer a more effective and reliable approach for quantifying the disaster impact. Ultimately, our research contributes to the advancement of disaster management and mitigation efforts.

\bibliographystyle{IEEEbib}
\bibliography{refer}

\end{document}